# Feature Selection and Dualities in Maximum Entropy Discrimination

**Tony Jebara**
MIT Media Lab
Massachusetts Institute of Technology
Cambridge, MA 02138

**Tommi Jaakkola**
MIT AI Lab
Massachusetts Institute of Technology
Cambridge, MA 02138

## Abstract

Incorporating feature selection into a classification or regression method often carries a number of advantages. In this paper we formalize feature selection specifically from a discriminative perspective of improving classification/regression accuracy. The feature selection method is developed as an extension to the recently proposed maximum entropy discrimination (MED) framework. We describe MED as a flexible (Bayesian) *regularization* approach that subsumes, e.g., support vector classification, regression and exponential family models. For brevity, we restrict ourselves primarily to feature selection in the context of linear classification/regression methods and demonstrate that the proposed approach indeed carries substantial improvements in practice. Moreover, we discuss and develop various extensions of feature selection, including the problem of dealing with example specific but unobserved degrees of freedom – alignments or invariants.

## 1 Introduction

Robust (discriminative) classification and regression methods have been successful in many areas ranging from image and document classification[7] to problems in biosequence analysis[5] and time series prediction[11]. Techniques such as Support vector machines[15], Gaussian process models[16], Boosting algorithms[1, 2], and more standard but related statistical methods such as logistic regression, are all robust against errors in structural assumptions. This property arises from a precise match between the training objective and the criterion by which the methods are subsequently evaluated.

Probabilistic (generative) models such as graphical models offer complementary advantages in classification or regression tasks such as the ability to deal effectively with uncertain or incomplete examples. Several approaches have been recently proposed for combining the generative and discriminative methods, including [4, 6, 14]. We provide an additional point of contact in the current paper.

The focus of this paper is on feature selection. The feature selection problem may involve finding the structure of a graphical model (as in [12]) or identifying a set of components of the input examples that are relevant for a classification task. More generally, feature selection can be viewed as a problem of setting discrete structural parameters associated with a specific classification or regression method. We subscribe here to the view that feature selection is not merely for reducing the computational load associated with a high dimensional classification or regression problem but can be tailored primarily to improve prediction accuracy (cf. [9]). This perspective excludes a number of otherwise useful feature selection approaches such as any *filtering* method that operates independently from the classification task/method at hand. Linear classifiers, for example, impose strict constraints about the type of features that are at all useful. Such constraints should be included in the objective function governing the feature selection process.

The form of feature selection we develop in this paper results in a type of feature weighting. Each feature or structural parameter is associated with a probability value. The feature selection process translates into estimating the most discriminative probability distribution over the structural parameters. Irrelevant features quickly receive low albeit non-zero probabilities of being selected. We emphasize that the feature selection is carried out jointly and discriminatively together with the estimation of the specific classification or regression method. This type of feature selection is, perhaps surprisingly, most beneficial when the number of training examples is relatively small compared to their dimensionality.

The paper is organized as follows. We begin by motivating the discriminative maximum entropy framework from the point of view of regularization theory. We then explicate how to solve classification and regression problems in the context of maximum entropy formalism and, subsequently, extend these ideas to feature selection by incorporating discrete structural parameters. Finally, we expose some future directions and problems.

## 2 Regularization framework and Maximum entropy

We begin by motivating the maximum entropy framework from the perspective of regularization theory. A reader interested primarily in feature selection and who may already be familiar with the maximum entropy framework may wish to skip this section except definition 1.

For simplicity, we will focus on binary classification; the extension to multi-class classification and regression problems is discussed later in the paper. Given a set of training examples $\{X_1, \ldots, X_T\}$ and the corresponding binary ($\pm 1$) labels $\{y_1, \ldots, y_T\}$, we seek to minimize some measure of classification error or loss within a chosen parametric family of decision boundaries such as linear. The decision boundaries are expressed in terms of *discriminant functions*, $\mathcal{L}(X; \Theta)$, the sign of which determines the predicted label.

We consider a specific class of loss functions, those that depend on the parameters $\Theta$ only through what is known as the *classification margin*. The margin, defined as $y_t \mathcal{L}(X_t; \Theta)$, is large and positive whenever the label $y_t$ agrees with the real valued prediction $\mathcal{L}(X_t; \Theta)$. We assume that the loss function, $L : \mathcal{R} \to \mathcal{R}$, is a *non-increasing* and *convex* function of the margin. Thus a larger margin accompanies a smaller loss. Many loss functions for classification problems are indeed of this type.

Given this class of margin loss functions $L(\cdot)$, we can define a regularization method for classification. Given a *convex* regularization penalty $R(\Theta)$ (typically the squared Euclidean norm), we estimate the parameters $\Theta$ by minimizing a combination of the empirical loss and the regularization penalty

$$J(\Theta) = \sum_t L(y_t \mathcal{L}(X_t; \Theta)) + R(\Theta)$$

The resulting $\hat{\Theta}$ can be subsequently used in the decision rule $y = \text{sign}\left(\mathcal{L}(X; \hat{\Theta})\right)$ to classify yet unseen examples.

Any regularization approach of this form admits a simple alternative description in terms of classification constraints. Given a convex non-increasing margin loss function $L(\cdot)$ as before, we can cast the minimization problem above as follows: minimize $R(\Theta) + \sum_t L(\gamma_t)$ with respect to $\Theta$ and the margin parameters $\gamma = [\gamma_1, \ldots, \gamma_T]$ subject to the classification constraints $y_t \mathcal{L}(X_t; \Theta) - \gamma_t \geq 0$, $\forall t$.

The maximum entropy framework proposed in [3] generalizes and clarifies this formulation in several respects. For example, we no longer find a fixed setting of the parameters $\Theta$ but a distribution over them. This generalization facilitates a number of extensions of the basic approach including feature selection described in this paper. The choice of the loss function (penalties for violating the margin constraints) also admits a more principled solution. We quote here a slightly rewritten (MED) formulation:

**Definition 1** *We find $P(\Theta, \gamma)$ over the parameters $\Theta$ and the margin variables $\gamma = [\gamma_1, \ldots, \gamma_T]$ that minimizes $KL(P_\Theta \| P_\Theta^0) + \sum_t KL(P_{\gamma_t} \| P_{\gamma_t}^0)$ subject to $\int P(\Theta, \gamma) [y_t \mathcal{L}(X_t, \Theta) - \gamma_t] d\Theta d\gamma \geq 0 \ \forall t$. Here $P_\Theta^0$ and $P_\gamma^0$ are the prior distributions over the parameters and the margin variables, respectively. The resulting decision rule is given by $\hat{y} = \text{sign}(\int P(\Theta) \mathcal{L}(X, \Theta) d\Theta)$.*

Note that in the above definition, we have relaxed the classification constraints into averaged constraints that are less restrictive in the sense that they need not hold for any specific parameter/margin value. Second, the regularization penalty (the analog of $R(\Theta)$) and the margin penalties (the analogs of $L(\gamma_t)$) are now measured on a common *scale*, i.e., in terms of KL-divergences. The common scale puts the inherent trade-off between these penalties on a more sound footing. Third, after specifying a prior distribution over the margin variables, we have fully specified the margin penalties: $KL(P_{\gamma_t} \| P_{\gamma_t}^0)$. This contributes a different perspective to the choice of the margin penalties.

Our probabilistic extension also admits an information theoretic interpretation. The method now minimizes the number of bits we have to extract from the training examples so as to satisfy the classification constraints. In this interpretation, the solution $P^*(\Theta, \gamma)$ is treated as the posterior distribution given the data. Under certain conditions on the prior $P^0(\Theta) P^0(\gamma)$, the expected penalty (the quantity being minimized) reduces to the mutual information between the data and the parameters. A more technical argument will be given in a longer version of the paper.

We could transform the maximum entropy formulation back into the regularization form and explicate the resulting loss functions and regularization penalties. Expressing the problem in terms of classification constraints seems, however, more flexible in a probabilistic context.

### 2.1 Solution

The solution to the MED classification problem in Definition 1 is directly solvable using a classical result from maximum entropy:

**Theorem 1** *The solution to the MED problem has the following general form (cf. Cover and Thomas 1996):*

$$P(\Theta, \gamma) = \frac{1}{Z(\lambda)} P_0(\Theta, \gamma) e^{\sum_t \lambda_t [y_t \mathcal{L}(X_t | \Theta) - \gamma_t]}$$

where $Z(\lambda)$ is the normalization constant (partition function) and $\lambda = \{\lambda_1, \ldots, \lambda_T\}$ defines a set of non-negative Lagrange multipliers, one per classification constraint. $\lambda$ are set by finding the unique maximum of the jointly concave objective function

$$J(\lambda) = -\log Z(\lambda) \tag{1}$$

Unfortunately, integrals are required to compute the log-partition function which may not always be analytically solvable. Furthermore, evaluation of the decision rule also requires an integral followed by a sign operation which may not be feasible for arbitrary choices of the priors and discriminant functions. However, it is generally true that if the discriminant arises from the ratio of two generative models[1] in the exponential family and the prior over the model is from the conjugate of that exponential family member, then the computations are tractable (see Appendix). In these cases, the discriminant function is:

$$\mathcal{L}(X;\Theta) = \log \frac{P(X|\theta_+)}{P(X|\theta_-)} + b \tag{2}$$

Here, b is a bias term that can be considered as a log-ratio of prior class probabilities. The variables $\{\theta_+, \theta_-\}$ are parameters and structures for the generative models in the exponential family for the positive and negative class respectively. Therefore, classification using linear decisions, multinomials, Gaussians, Poisson, tree-structured graphs and other exponential family members are all accommodated. Generative models outside the exponential family may still be accommodated although approximations such as meanfield might be necessary.

Once the concave objective function is given (possibly with a convex hull of constraints), optimization towards the unique maximum can be done with a variety of techniques. Typically, we utilize a randomized axis-parallel line search (i.e. searching with Brent's method) in each of the directions of $\lambda$.

### 2.2 Dual priors and penalty functions

Expanding the definition of the objective function in Theorem 1, we obtain the following log-partition to minimize in $\lambda$ with constraints on the variables (i.e. positivity among other possibilities):

$$\begin{aligned}\log Z &= \log\left(\int P_0(\Theta)e^{\sum_t \lambda_t y_t \mathcal{L}(X_t|\Theta)}d\Theta\right) \\ &+ \sum_t \log\left(\int P_0(\gamma_t)e^{-\lambda_t \gamma_t}d\gamma_t\right) \\ &= \log Z_\Theta(\lambda) + \sum_t \log Z_{\gamma_t}(\lambda_t)\end{aligned}$$

---

[1] Note, here we shall use the term generative model to mean a distribution over data whose parameters and structure are estimated without necessarily resorting to traditional Bayesian approaches.

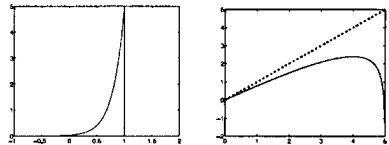

Figure 1: Margin prior distribution (left) and associated penalty function (right).

Note the factorization of $P(\Theta, \gamma)$ into $P(\Theta)\Pi_t P(\gamma_t)$ due to the original factorization in the prior $P_0$. This objective function is also similar to the definition of $J(\Theta)$ in the regularization approach. We now have a direct way of finding penalty terms $\log Z_{\gamma_t}(\lambda_t)$ from margin priors $P_0(\gamma_t)$ and vice-versa. Thus, there is a dual relationship between defining an objective function and penalty terms and defining a prior distribution over parameters and prior distribution over margins.

For instance, consider the prior margin distribution $P(\gamma) = \Pi_t P(\gamma_t)$ where

$$P(\gamma_t) = ce^{-c(1-\gamma_t)}, \gamma_t \leq 1 \tag{3}$$

Integrating, we get the penalty function (Figure 1):

$$\begin{aligned}\log Z_{\gamma_t}(\lambda_t) &= \log \int_{\gamma_t=-\infty}^{1} ce^{-c(1-\gamma_t)}e^{-\lambda_t \gamma_t}d\gamma_t \\ &= \lambda_t + \log(1 - \lambda_t/c)\end{aligned}$$

Figure 1 shows the above prior and its associated penalty term.

### 2.3 SVM Classification

Using the MED formulation and assuming a linear discriminant function with a Gaussian prior on the weights produces support vector machines:

**Theorem 2** *Assuming $\mathcal{L}(X;\Theta) = \theta^T X + b$ and $P_0(\Theta,\gamma) = P_0(\theta)P_0(b)P_0(\gamma)$ where $P_0(\theta)$ is $N(0,I)$, $P_0(b)$ approaches a non-informative prior, and $P_0(\gamma)$ is given by $P_0(\gamma_t)$ as in Equation 3 then the Lagrange multipliers $\lambda$ are obtained by maximizing $J(\lambda)$ subject to $0 \leq \lambda_t \leq c$ and $\sum_t \lambda_t y_t = 0$, where*

$$J(\lambda) = \sum_t [\lambda_t + \log(1 - \lambda_t/c)] - \frac{1}{2}\sum_{t,t'}\lambda_t \lambda_{t'} y_t y_{t'}(X_t^T X_{t'})$$

The only difference between our $J(\lambda)$ and the (dual) optimization problem for SVMs is the additional potential term $\log(1 - \lambda_t/c)$ which acts as a barrier function preventing the $\lambda$ values from growing beyond $c$. This highlights the effect of the different missclassification penalties. In the separable case, letting $c \to \infty$, the two methods coincide. The decision rules are formally identical.

### 2.4 Probability Density Classification

Other discriminant functions can be accommodated, including likelihood ratios of probability models. This permits the concepts of large margin and support vectors to operate in a generative model setting. For instance, one could consider the discriminant that arises from the likelihood ratio of two Gaussians: $\mathcal{L}(X;\Theta) = \log \mathcal{N}(\mu_1, \Sigma_1) - \log \mathcal{N}(\mu_2, \Sigma_2) + b$ or the likelihood ratio of two tree-structures models. This and other discriminative classifications using non-SVM models are detailed in [3]. Also, refer to the Appendix in this paper for derivations related to general exponential family densities.

It is straightforward to perform multi-class discriminative density estimation by adding extra classification constraints. The binary case merely requires T inequalities of the form: $y_t \mathcal{L}(X_t; \Theta) - \gamma_t \geq 0, \forall t$. In a multi-class setting, constraints are needed for all pairwise log-likelihood ratios. In other words, in a 3 class problem $(A, B, C)$, with 3 models $(\theta_A, \theta_B, \theta_C)$, if $y_t = A$, the log-likelihood of model $\theta_A$ must dominate. In other words, we have the following two classification constraints:

$$\int P(\Theta, \gamma)[\log \frac{P(X_t|\theta_A)}{P(X_t|\theta_B)} + b_{AB} - \gamma] d\Theta d\gamma \geq 0$$

$$\int P(\Theta, \gamma)[\log \frac{P(X_t|\theta_A)}{P(X_t|\theta_C)} + b_{AC} - \gamma] d\Theta d\gamma \geq 0$$

## 3 MED Regression

The MED formalism is not restricted to classification. It can also accommodate other tasks such as anomaly detection [3]. Here, we present its extension to the regression (or function approximation) case using the approach and nomenclature in [13]. Dual sided constraints are imposed on the output such that an interval called an $\epsilon$-tube around the function is described [2]. Suppose training input examples $\{X_1, \ldots, X_T\}$ are given with their corresponding output values as continuous scalars $\{y_1, \ldots, y_T\}$. We wish to solve for a distribution of parameters of a discriminative regression function as well as margin variables:

**Theorem 3** *The maximum entropy discrimination regression problem can be cast as follows:*

*Find $P(\Theta, \gamma)$ that minimizes $KL(P\|P_0)$ subject to the constraints:*

$$\int P(\Theta, \gamma) [y_t - \mathcal{L}(X_t; \Theta) + \gamma_t] d\Theta d\gamma \geq 0, \quad t = 1..T$$
$$\int P(\Theta, \gamma) [\gamma'_t - y_t + \mathcal{L}(X_t; \Theta)] d\Theta d\gamma \geq 0, \quad t = 1..T$$

*where $\mathcal{L}(X_t; \Theta)$ is a discriminant function and $P_0$ is a prior distribution over models and margins. The de-*

---
[2] An $\epsilon$-tube (as in the SVM literature) is a region of insensitivity in the loss function which only penalizes approximation errors which deviate by more than $\epsilon$ from the data.

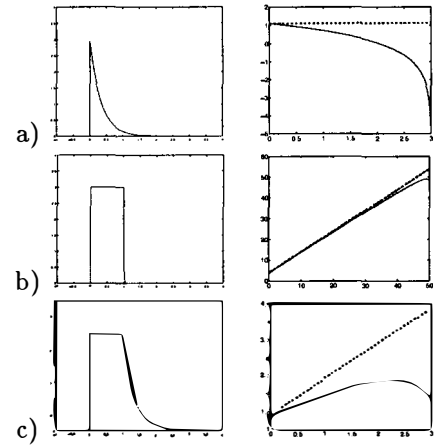

Figure 2: Margin prior distribution (left) and associated penalty function (right).

*cision rule is given by $\hat{y} = \int P(\Theta) \mathcal{L}(X;\Theta) d\Theta$. The solution is given by:*

$$P(\Theta, \gamma) = \frac{1}{Z(\lambda)} P_0(\Theta, \gamma) \frac{e^{\sum_t \lambda_t [y_t - \mathcal{L}(X_t|\Theta) + \gamma_t]}}{e^{\sum_t \lambda'_t [y_t - \mathcal{L}(X_t|\Theta) - \gamma'_t]}}$$

*where the objective function is again* $-\log Z(\lambda)$.

Typically, we have the following prior for $\gamma$ which differs from the classification case due to the additive role of the output $y_t$ (versus multiplicative) and the two-sided constraints.

$$P(\gamma_t) \propto \left\{ \begin{array}{ll} 1 & \text{if} \quad 0 \leq \gamma_t \leq \epsilon \\ e^{c(\epsilon - \gamma_t)} & \text{if} \quad \gamma_t > \epsilon \end{array} \right\} \quad (4)$$

Integrating, we obtain:

$$\log Z_{\gamma_t}(\lambda_t) = \log \int_0^\epsilon e^{\lambda_t \gamma_t} d\gamma_t + \int_\epsilon^\infty e^{c(\epsilon - \gamma_t)} e^{\lambda_t \gamma_t} d\gamma_t$$

$$\log Z_{\gamma_t}(\lambda_t) = \epsilon \lambda_t - \log(\lambda_t) + \log\left(1 - e^{-\lambda_t \epsilon} + \frac{\lambda_t}{c - \lambda_t}\right)$$

Figure 2 shows the above prior and its associated penalty terms under different settings of $c$ and $\epsilon$. Varying $\epsilon$ effectively modifies the thickness of the $\epsilon$-tube around the function. Furthermore, $c$ varies the robustness to outliers by tolerating violations of the $\epsilon$-tube.

### 3.1 SVM Regression

If we assume a linear discriminant function for $\mathcal{L}$ (or linear decision after a Kernel), the MED formulation generates the same objective function that arises in SVM regression [13]:

**Theorem 4** *Assuming $\mathcal{L}(X; \Theta) = \theta^T X + b$ and $P_0(\Theta, \gamma) = P_0(\theta) P_0(b) P_0(\gamma)$ where $P_0(\theta)$ is $N(0, I)$, $P_0(b)$ approaches a non-informative prior, and $P_0(\gamma)$ is given by Equation 4 then the Lagrange multipliers $\lambda$ are obtained by maximizing $J(\lambda)$ subject to $0 \leq \lambda_t \leq c$, $0 \leq \lambda'_t \leq c$ and $\sum_t \lambda_t = \sum_t \lambda'_t$, where*

$$J(\lambda) = \sum_t y_t (\lambda'_t - \lambda_t) - \epsilon \sum_t (\lambda_t + \lambda'_t)$$

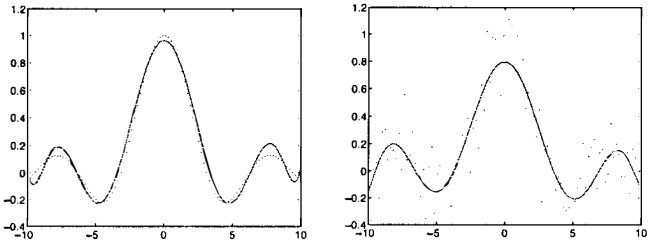

Figure 3: MED approximation to the sinc function: noise-free case (left) and with Gaussian noise (right).

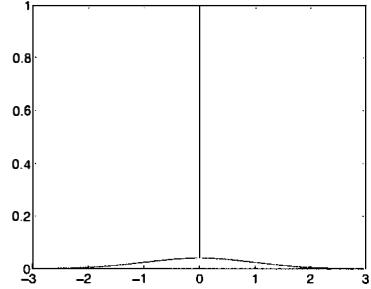

Figure 4: The prior distribution over $\theta_i s_i$.

$$+ \sum_t \log(\lambda_t) - \log\left(1 - e^{-\lambda_t \epsilon} + \frac{\lambda_t}{c - \lambda_t}\right)$$

$$+ \sum_t \log(\lambda'_t) - \log\left(1 - e^{-\lambda'_t \epsilon} + \frac{\lambda'_t}{c - \lambda'_t}\right)$$

$$- \frac{1}{2} \sum_{t,t'} (\lambda_t - \lambda'_t)(\lambda_{t'} - \lambda'_{t'})(X_t^T X_{t'})$$

As can be seen (and more so as $c \to \infty$), the objective becomes very similar to the one in SVM regression. There are some additional penalty functions (all the logarithmic terms) which can be considered as barrier functions in the optimization to maintain the constraints.

To illustrate the regression, we approximate the sinc function, a popular example in the SVM literature. Here, we sampled 100 points from the sinc$(x) = |x|^{-1} \sin|x|$ within the interval [-10,10]. We also considered a noisy version of the sinc function where Gaussian additive noise of standard deviation 0.2 was added to the output. Figure 3 shows the resulting function approximation which is very similar to the SVM case. The Kernel applied was an 8th order polynomial [3].

## 4 Feature selection in classification

We now extend the formulations to accomodate feature selection. We begin with the classification case. For simplicity, consider only linear classifiers and parameterize the discriminant function as follows

$$\mathcal{L}(X; \Theta) = \sum_{i=1}^n \theta_i s_i X_i + \theta_0$$

where $\Theta = \{\theta_0, \ldots, \theta_n, s_1, \ldots, s_n\}$ now also contains binary structural parameters $s_i \in \{0, 1\}$. These either select or exclude a particular component of the input vector $X$. Recall that there is no inherent difference between discrete and continuous variables in the MED formalism since we are primarily dealing with only distributions over such parameters [3].

To completely specify the learning method in this context, we have to define a prior distribution over the parameters $\Theta$ as well as over the margin variables $\gamma$. For the latter, we use the prior described in Eq. (3). The choice of the prior $P_0(\Theta)$ is critical as it determines the effect of the discrete parameters $s$. For example, assigning a larger prior probability for $s_i = 1, \forall i$ simply reduces the problem to the standard formulation discussed earlier. We provide here one reasonable choice:

$$P_0(\Theta) = P_{0,\theta_0}(\theta_0) P_{0,\theta}(\theta) \prod_{i=1}^n P_{s,0}(s_i)$$

where $P_{0,\theta_0}$ is an uninformative prior[4], $P_{\theta,0}(\theta) = \mathcal{N}(0, I)$, and

$$P_{s,0}(s_i) = p_0^{s_i} (1 - p_0)^{1-s_i}$$

where $p_0$ controls the overall prior probability of including a feature. This prior should be viewed in terms of the distribution that it defines over $\theta_i s_i$. The figure below illustrates this for one component.

### 4.1 The log-partition function

Having defined the prior distribution over the parameters in the MED formalism, it remains to evaluate the partition function (cf. Eq. (1)). Again we first remove the effect of the bias variable and obtain the additional constraint[5] $\sum_t \lambda_t y_t = 0$ on the Lagrange multipliers associated with the classification constraints. Omitting the straightforward algebra, we obtain

$$J(\lambda) = -\log Z(\lambda)$$

---

[3] A Kernel implicitly transforms the input data by modifying the dot-product between data vectors $k(X_t, X'_t) = \langle \Phi(X_t), \Phi(X'_t) \rangle$. This can also be done by explicitly remapping the data via the transformation $\Phi(X_t)$ and using the conventional dot-product. This permits non-linear classification and regression using the basic linear SVM machinery. For example, an $m$-th order polynomial expansion replaces a vector $X_t$ by $\Phi(X_t) = [X_t; X_t^2; \ldots X_t^m]$.

[4] Or a zero mean Gaussian prior with a sufficiently large variance.

[5] Alternatively, if a broad Gaussian prior ($\sigma \gg 1$) is used for the bias term, we would end up with a quadratic penalty term $-\frac{\sigma^2}{2} \sum_t \lambda_t y_t$ in the objective function $J(\lambda)$ but without the additional constraint $\sum_t \lambda_t y_t = 0$. This soft constraint often simplifies the optimization of $J(\lambda)$ and for sufficiently large $\sigma$ has no effect on the solution.

$$= \sum_t [\lambda_t + \log(1 - \lambda_t/c)]$$

$$- \sum_{i=1}^{n} \log\left[1 - p_0 + p_0 e^{\frac{1}{2}(\sum_t \lambda_t y_t X_{t,i})^2}\right]$$

which we maximize subject to $\sum_t \lambda_t y_t = 0$.

This closed form expression for $\log Z(\lambda)$ allows us to study further the properties of the resulting maximum entropy distribution over $\theta_i s_i$. The mean of this distribution is readily found by observing that

$$\frac{\partial \log Z(\lambda_t)}{\partial \lambda_t} = E_P\{ y_t \sum_i \theta_i s_i X_{t,i} - \gamma_t \}$$

$$= y_t \sum_i E_P\{\theta_i s_i\} X_{t,i} - E_P\{\gamma_t\}$$

$$= y_t \sum_i P_i(\sum_t \lambda_{t'} y_{t'} X_{t',i}) X_{t,i} - (1 - \frac{1}{c - \lambda_t})$$

where the expectations are with respect to the maximum entropy distribution. (note that the average over the bias term is missing since we did not include it in the definition of the partition function $Z(\lambda)$). Here $P_i$ is defined as

$$P_i = \text{Logistic}\left[(\sum_{t'} \lambda_t y_{t'} X_{t',i})^2 + \log \frac{p_0}{1 - p_0}\right]$$

We denote $W_i = \sum_{t'} \lambda_t y_{t'} X_{t',i}$, which is formally identical to the average $E_P\{\theta_i\}$ in the absence of the selection variables $s_i$ (i.e., without feature selection). In our case,

$$E_P\{\theta_i s_i\} = \text{Logistic}\left[W_i^2 + \log \frac{p_0}{1 - p_0}\right] W_i$$

We may now understand the effect of the discrete selection variables by comparing the functional form of the above average with $W_i$ as $W_i$ is varied.

The figure below illustrates $P_i(W_i) W_i$ and $W_i$ for positive values of $W_i$. The effect of the feature selection is clearly seen in terms of the rapid non-linear decay of the effective coefficient $P_i(W_i) W_i$ with decreasing $W_i$. The two graphs merge for larger values of $W_i$ corresponding to the setting $s_i = 1$. The location where the selection takes place depends on the prior probability of $p_0$, and happens around

$$W_i^* = \pm \sqrt{\log \frac{1 - p_0}{p_0}}$$

In Figure 5, $p_0 = 0.01$.

### 4.2 Experimental results

We tested our linear feature selection method on a DNA splice site recognition problem, where the problem is to distinguish true and spurious splice sites. The examples were fixed length DNA sequences (length 25) that we binary encoded (4 bit translation of $\{A, C, T, G\}$) into a vector of 100 binary components.

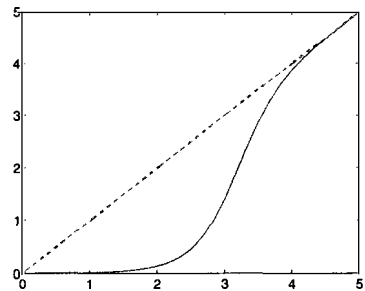

Figure 5: The behavior of the linear coefficients with and without feature selection. In feature selection, smaller coefficients have greatly diminished effects (solid line).

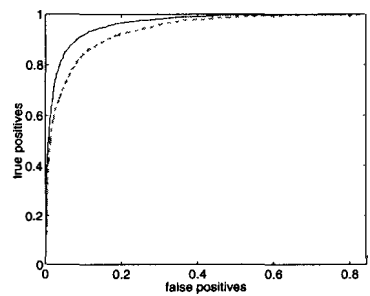

Figure 6: ROC curves on the splice site problem with feature selection $p_0 = 0.00001$ (solid line) and without $p_0 = 0.99999$ (dashed line).

The training set consisted of 500 examples and the independent test set contained 4724 examples. Figure 6 illustrates the benefit arising from the feature selection approach.

In order to verify that the feature selection indeed greatly reduces the effective number of components, we computed the empirical cumulative distribution functions of the magnitudes of the resulting coefficients $\hat{P}(|\tilde{W}| < x)$ as a function of $x$ based on the 100 components. In the feature selection context, the linear coefficients are $\tilde{W}_i = E_P\{\theta_i s_i\}$, $i = 1, \ldots, 100$ and $\tilde{W}_i = E_P\{\theta_i\}$ when no feature selection is used. These coefficients appear in the decision rules in the two cases and thus provide a meaningful comparison. Figure 7 indicates that most of the weights resulting from the feature selection algorithm are indeed small enough to be neglected.

Since the complexity of the feature selection algorithm scales only linearly in the number of original features (components), we can also use quadratic componentwise expansions of the examples as the input vectors. Figure 7 below shows that the benefit from the feature selection algorithm does not degrade as the number of features increases (in this case $\approx 5000$).

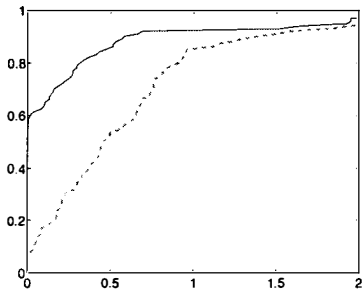

Figure 7: Cumulative distribution functions for the resulting effective linear coefficients with feature selection (solid line) and without (dashed line).

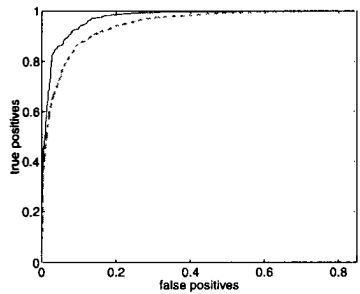

Figure 8: ROC curves corresponding to a quadratic expansion of the features with feature selection $p_0 = 0.00001$ (solid line) and without $p_0 = 0.99999$ (dashed line).

| Linear Model Estimator | $\epsilon$-sensitive linear loss |
|---|---|
| Least-Squares Fit | 1.7584 |
| MED $p_0 = 0.99999$ | 1.7529 |
| MED $p_0 = 0.1$ | 1.6894 |
| MED $p_0 = 0.001$ | 1.5377 |
| MED $p_0 = 0.00001$ | 1.4808 |

Table 1: Prediction Test Results on Boston Housing Data. Note, due to data rescaling, only the relative quantities here are meaningful.

## 5 Feature selection in regression

Feature selection can also be advantageous in the regression case where a map is learned from inputs to scalar outputs. Since some input features might be irrelevant (especially after a Kernel expansion), we again employ an aggressive pruning approach by adding a "switch" $(s_i)$ on the parameters as before. The prior is given by $P_0(s_i) = p_0^{s_i}(1-p_0)^{1-s_i}$ where lower values of $p_0$ encourage further sparsification. This prior is in addition to the Gaussian prior on the parameters $(\Theta_i)$ which does not have quite the same sparsification properties.

The previous derivation for feature selection can also be applied in a regression context. The same priors are used except that the prior over margins is swapped with the one in Equation 4. Also, we shall include the estimation of the bias in this case, where we have a Gaussian prior: $P_0(b) = \mathcal{N}(0, \sigma)$. This replaces the hard constraint that $\sum_t \lambda_t = \sum_t \lambda'_t$ with a soft quadratic penalty, making computations simpler. After some straightforward algebraic manipulations, we obtain the following form for the objective function:

$$\begin{aligned} J(\lambda) = & \sum_t y_t(\lambda'_t - \lambda_t) - \epsilon \sum_t (\lambda_t + \lambda'_t) \\ & - \tfrac{1}{2}\sigma(\sum_t \lambda_t - \lambda'_t)^2 \\ & + \sum_t \log(\lambda_t) - \log\left(1 - e^{-\lambda_t \epsilon} + \tfrac{\lambda_t}{c - \lambda_t}\right) \\ & + \sum_t \log(\lambda'_t) - \log\left(1 - e^{-\lambda'_t \epsilon} + \tfrac{\lambda'_t}{c - \lambda'_t}\right) \\ & - \sum_i \log\left(1 - p_0 + p_0 e^{\tfrac{1}{2}[\sum_t (\lambda_t - \lambda'_t)X_{t,i}]^2}\right) \end{aligned}$$

This objective function is optimized over $(\lambda_t, \lambda'_t)$ and by concavity has a unique maximum. The optimization over Lagrange multipliers controls optimization of the densities of the model parameter settings $P(\Theta)$ as well as the switch settings $P(s)$. Thus, there is a *joint* discriminative optimization over feature selection and parameter settings.

### 5.1 Experimental Results

Below, we evaluate the feature selection based regression (or Support Feature Machine, in principle) on a popular benchmark dataset, the 'Boston housing' problem from the UCI repository. A total of 13 features (all treated continuously) are given to predict a scalar output (the median value of owner-occupied homes in thousands of dollars). To evaluate the dataset, we utilized both a linear regression and a 2nd order polynomial regression by applying a Kernel expansion to the input. The dataset is split into 481 training samples and 25 testing samples (as in [14]).

Table 1 indicates that feature selection (decreasing $p_0$) generally improves the discriminative power of the regression. Here, the $\epsilon$-sensitive linear loss functions (typical in the SVM literature) shows improvements with further feature selection. Just as sparseness in the number of vectors helps generalization, sparseness in the number of features is advantageous as well. Here, there is a total of 104 input features after the 2nd order polynomial Kernel expansion. However, not all have the same discriminative power and pruning is beneficial.

For the 3 trial settings of the sparsification level prior ($p_0 = 0.99999, p_0 = 0.001, p_0 = 0.00001$), we again analyze the cumulative density function of the resulting linear coefficients $\hat{P}(|\bar{W}| < x)$ as a function of $x$ based on the features from the Kernel expansion. Figure 9 clearly indicates that the magnitudes of the coefficients are reduced as the sparsification prior is increased.

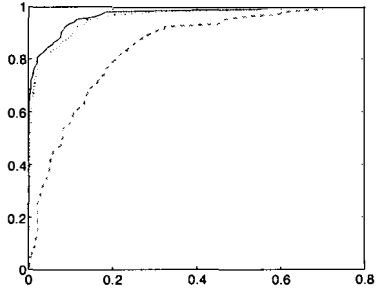

Figure 9: Cumulative distribution functions for the linear regression coefficients under various levels of sparsification. Dashed line: $p_0 = 0.99999$, dotted line: $p_0 = 0.001$ and solid line: $p_0 = 0.00001$.

| Linear Model Estimator | $\epsilon$-sensitive linear loss |
|---|---|
| Least-Squares Fit | 3.609e+03 |
| MED $p_0 = 0.00001$ | 1.6734e+03 |

Table 2: Prediction Test Results on Gene Expression Level Data.

The MED regression was also used to predict gene expression levels using data from "Systematic variation in gene expression in human cancer cell lines", by D. Ross et. al. Here, log-ratios ($\log(RAT2n)$) of gene expression levels were to be predicted for a Renal Cancer cell-line from measurements of each gene's expression levels across different cell-lines and cancer-types. Input data forms a 67-dimensional vector while output is a 1-dimensional scalar gene expression level. Training set size was limited to 50 examples and testing was over 3951 examples. The table below summarizes the results. Here, an $\epsilon = 0.2$ was used along with $c = 10$ for the MED approach. This indicates that the feature selection is particularly helpful in sparse training situations.

## 6  Discriminative feature selection in generative models

As mentioned earlier, the MED framework is not restricted to discriminant functions that are linear or non-probabilistic. For instance, we can consider the use of feature selection in a generative model-based classifier. One simple case is the discriminant formed from the ratio of two identity-covariance Gaussians. Parameters $\Theta$ are $(\mu, \nu)$ for the means of the $y = +1$ and $y = -1$ classes respectively and the discriminant is $\mathcal{L}(X; \Theta) = \log \mathcal{N}(\mu, I) - \log \mathcal{N}(\nu, I) + b$. As before, we insert switches ($s_i$ and $r_i$) to turn off certain components of each of the Gaussians giving us:

$$\mathcal{L}(X; \Theta) = \sum_i s_i (X_i - \mu_i)^2 - \sum_i r_i (X_i - \nu_i)^2 + b$$

This discriminant then uses the similar priors to the ones previously introduced for feature selection in a linear classifier. It is straightforward to integrate (and *sum* over discrete $s_i$ and $r_i$) with these priors (shown below and in Equation 3) to get an analytic concave objective function $J(\lambda)$:

$$P_0(\mu) = \mathcal{N}(0, I) \qquad P_0(\nu) = \mathcal{N}(0, I)$$
$$P_0(s_i) = p_0^{s_i}(1-p_0)^{1-s_i} \qquad P_0(r_i) = p_0^{r_i}(1-p_0)^{1-r_i}$$

In short, optimizing the feature selection and means for these generative models jointly will produce degenerate Gaussians which are of smaller dimensionality than the original feature space. Such a feature selection process could be applied to many density models in principle but computations may require mean-field or other approximations to become tractable.

## 7  Example-specific features, latent variables and transformations

Another extension of the MED framework concerns feature selection with example-specific degrees of freedom such as invariant transformations or alignments (the idea and the problem formulation resemble those proposed in [10]). For example, assume for each input vector in $\{X_1, \ldots, X_T\}$ we are given not only a binary class label in $\{y_1, \ldots, y_T\}$ but also a hidden *transformation* variable in $\{U_1, \ldots, U_T\}$. The transformation variable modifies the input space to generate a different $\hat{X} = \mathcal{T}(X, U)$. The transformation $U_t$ associated with each data point is, however, unknown with some prior probability $P_0(U_t)$. For example, the discriminant function could be defined as $\mathcal{L}(X_t, \Theta) = \Theta^T(X_t - U_t \vec{1}) + b$, where the scalar $U_t$ represents a translation along $\vec{1}$. More generally, the presence of the latent transformation variables $U$ encode invariants. The MED solution would then be given by:

$$P(\Theta, U, \gamma) = \frac{1}{Z(\lambda)} P_0(\Theta, U, \gamma) e^{\sum_t \lambda_t [y_t \mathcal{L}(X_t - U_t \vec{1} | \Theta) - \gamma_t]}$$

In this discriminative formulation, the solution can be obtained only in a *transductive* sense [15]. In other words, bias for selecting the latent transformations comes from the preference towards large margin classification. Any set of new examples to be classified possess independent transformation variables. They must be included with the training examples as unlabeled examples to exploit the bias. The solution is obtained similarly to the treatment of ordinary unlabeled examples in [3]. More specifically, we can make use of a mean-field approximation to iteratively optimize the relevant distributions. First, we hypothesize a marginal distribution over the transformation variables (such as the prior), fix these distributions and update $P(\Theta)$ independently. The resulting $P(\Theta)$ would be in turn held constant and the $P(U)$ updated and so on. The convergence of such alternating optimization is guaranteed as in [3].

As an example, consider transformations that correspond to warping of a temporal signal. If $X$ is a time varying multi-dimensional signal, we could align it to a model such as a hidden Markov model. The HMM specification provides the ordinary parameters in this context while the hidden state sequence takes the role of the individual transformations. Further experiments relating to this will be made available at:

http://www.media.mit.edu/~jebara/med

## 8 Discussion

We have formalized feature selection as an extension of the maximum entropy discrimination formalism, a Bayesian regularization approach. The selection of features is carried out by finding the most discriminative probability distribution over the structural selection parameters or transformations corresponding to the features. Such calculations were shown to be feasible in the context of linear classification/regression methods and when the discriminant functions arise from log-likelihood ratios of class-conditional distributions in the exponential family. Our experimental results support the contention that discriminative feature selection indeed accompanies a substantial improvement in prediction accuracy. Finally, the feature selection formalism was further extended to cover unobserved degrees of freedom associated with individual examples such as invariances or alignments.

## A  Exponential Family

As mentioned in the text, discriminant functions that can be efficiently solved within the MED approach include log-likelihood ratios of the exponential family of distributions. This family subsumes a wide set of distributions and its members are characterized by the following form: $p(X|\theta) = \exp(A(X) + X^T\theta - K(\theta))$ for any convex $K$. Each family member has a conjugate prior distribution given by $p(\theta|\chi) = exp(\tilde{A}(\theta) + \theta^T\chi - \tilde{K}(\chi))$; here $\tilde{K}$ is also convex.

Whether or not a specific combination of a discriminant function and an associated prior over the parameters is feasible within the MED framework depends on whether we can evaluate the partition function (the objective function used for optimizing the Lagrange multipliers associated with the constraints). In general, these operations will require integrals over the associated parameter distributions. In particular, recall the partition function corresponding to the binary classification case (Section 2.2). Consider the integral over $\Theta$ in:

$$Z_\Theta(\lambda) = \int P_0(\Theta) e^{\sum_t \lambda_t y_t L(X_t|\Theta)} d\Theta$$

If we now separate out the parameters associated with the class-conditional densities as well as the bias term (i.e. $\theta_+, \theta_-, b$) and expand the discriminant function as a log-likelihood ratio, we obtain the following:

$$Z_\Theta = \int P_0(\theta_+)P_0(\theta_-)P_0(b) e^{\sum_t \lambda_t y_t [\log \frac{P(X|\theta_+)}{P(X|\theta_-)} + b]} d\Theta$$

which factorizes as $Z_\Theta = Z_{\theta_+} Z_{\theta_-} Z_b$. We can now substitute the exponential family forms for the class-conditional distributions and associated conjugate distributions for the priors. We assume that the prior is defined by specifying a value for $\chi$. It suffices here to show that we can obtain $Z_\theta^+$ in closed form. For simplicity, we drop the class identifier "+". The problem is now reduced to evaluating

$$Z_\theta(\lambda) = \int e^{\tilde{A}(\theta) + \theta^T\chi - \tilde{K}(\chi)}$$
$$\times e^{\sum_t \lambda_t y_t (A(X_t) + X_t^T\theta - K(\theta))} d\theta$$

We have shown earlier (see Theorem 2 or [3]) in the paper that a non-informative prior over the bias term $b$ leads to the constraint $\sum_t \lambda_t y_t = 0$. Making this assumption, we get

$$Z_\theta(\lambda) = e^{-\tilde{K}(\chi) + \sum_t \lambda_t y_t A(X_t)} \times$$
$$\int e^{\tilde{A}(\theta) + \theta^T(\chi + \sum_t \lambda_t y_t X_t)} d\theta$$
$$= e^{-\tilde{K}(\chi) + \sum_t \lambda_t y_t A(X_t)} \times e^{\tilde{K}(\chi + \sum_t \lambda_t y_t X_t)}$$

where the last evaluation is a property of the exponential family. The expressions for $A, \tilde{A}, K, \tilde{K}$ are known for specific distributions in the exponential family and can easily be used to complete the above evaluation, or realize the objective function (which is holds for any exponential-family distribution):

$$\log Z_\theta(\lambda) = \tilde{K}(\chi + \sum_t \lambda_t y_t X_t) + \sum_t \lambda_t y_t A(X_t) - \tilde{K}(\chi)$$

## B  Optimization & Bounded Quadratic Programming

The aforementioned MED approaches all employ a concave objective function $J(\lambda)$ with convex constraints. This is a powerful paradigm since it guarantees consistence convergence to unique solutions and is not sensitive to initialization conditions and local minima. Experiments are thus repeatable for the settings of the variables $(c, \epsilon, p_0, \sigma)$. The main computational requirement is an efficient way to maximize $J(\lambda)$.

One approach is to perform line searches in each $\lambda_t$ variable in an axis-parallel way. Due to the SVM-like structure, computations simplify if only one $\lambda_t$ variable is modified at a time. This approach works well in the classification case where there is only a single $\lambda_t$ per data point. However, in the regression case, the degrees of freedom double and a $\lambda_t$ and $\lambda_t'$ are available for each data point. This slows down convergence.

Alternatively, we can map the concave objective function to a quadratic programming problem (QP) by finding a variational quadratic lower bound on $J(\lambda)$. We can then iterate the bound computation with QP solutions and guarantee convergence to the global maximum. Recall, for example, the $J(\lambda)$ defined Equation 4. There are non-quadratic terms due to the log-potential functions as well as the last sum of logarithmic terms. The log-potential functions are not critical since the convex constraints subsume them. The only remaining dominant non-quadratic terms are thus those inside $\sum_i$, namely:

$$\begin{aligned} j_i(\lambda) &= -\log\left(1 - p_0 + p_0 e^{\frac{1}{2}[\sum_t (\lambda_t - \lambda'_t) X_{t,i}]^2}\right) \\ &= -\log\left(1 - p_0 + p_0 e^{\frac{1}{2}\lambda_t^T M \lambda}\right) \end{aligned}$$

Each of these can be lower bounded by the following expression which makes tangential contact at the current locus of optimization ($\tilde{\lambda}$) as follows:

$$j_i(\lambda) \geq \lambda^T (N + hM)\tilde{\lambda} - \frac{1}{2}\lambda^T (M + N)\lambda + const.$$

where

$$\begin{aligned} N &= \frac{1}{4}(M\tilde{\lambda})(M\tilde{\lambda})^T \\ h &= (1 - p0)/\left(1 - p0 + p0 e^{\frac{1}{2}\lambda_t^T M \lambda}\right) \end{aligned}$$

This approach requires a few iterations of QP to converge. Since subsequent QP iterations can reuse the previous step's solution as a seed, QP computations after the first are much faster. Thus, training is computationally efficient and converges in under 4X that of regular SVM QP solutions. The iterated bounded QP approach is recommended as a fast bootstrap for the axis-parallel search which can further optimize the true objective function subsequently (i.e. it fully considers the log-potential terms). On the other hand, QP may become intractable for very large data sets (the data matrix grows as the squared of the data set size) and there axis-parallel techniques alone would be preferable.